# TriP-LLM: A Tri-Branch Patch-wise Large Language Model Framework for Time-Series Anomaly Detection

**Yuan-Cheng Yu, Yen-Chieh Ouyang, Life Member, IEEE, Chun-An Lin**
Department of Electrical Engineering at National Chung Hsing University, Taichung 402202, Taiwan

Corresponding author: Yen-Chieh Ouyang (ycouyang@nchu.edu.tw)

**ABSTRACT** Time-series anomaly detection plays a central role across a wide range of application domains. With the increasing proliferation of the Internet of Things (IoT) and smart manufacturing, time-series data has dramatically increased in both scale and dimensionality. This growth has exposed the limitations of traditional statistical methods in handling the high heterogeneity and complexity of such data. Inspired by the recent success of large language models (LLMs) in multimodal tasks across language and vision domains, we propose a novel unsupervised anomaly detection framework: A Tri-Branch Patch-wise Large Language Model Framework for Time-Series Anomaly Detection (TriP-LLM). TriP-LLM integrates local and global temporal features through a triple-branch design comprising Patching, Selecting, and Global modules, to encode the input time-series into patch-wise representations, which are then processed by a frozen, pretrained LLM. A lightweight patch-wise decoder reconstructs the input, from which anomaly scores are derived. We evaluate TriP-LLM on several public benchmark datasets using PATE, a recently proposed threshold-free evaluation metric, and conduct all comparisons within a unified open-source framework to ensure fairness. Experimental results show that TriP-LLM consistently outperforms recent state-of-the-art (SOTA) methods across all datasets, demonstrating strong detection capabilities. Furthermore, through extensive ablation studies, we verify the substantial contribution of the LLM to the overall architecture. Compared to LLM-based approaches using Channel Independence (CI) patch processing, TriP-LLM achieves significantly lower memory consumption, making it more suitable for GPU memory-constrained environments. All code and model checkpoints of TriP-LLM are publicly available on https://github.com/YYZStart/TriP-LLM.git

## I. INTRODUCTION

Over the past few decades, time-series analysis has been a cornerstone of data science and artificial intelligence (AI) research. Across all kinds of data, such as stock market quotes, sensor data from the Internet of Things (IoT), network traffic, and event logs from smart manufacturing systems are all naturally recorded as time-series sequences [1]. Classical statistical and traditional machine learning (ML) models, such as Autoregressive Integrated Moving Average (ARIMA) [2] and Isolation Forest (IF) [3] and Support Vector Machines (SVMs) [4], have been widely used to make reliable forecasts and detect anomalies in these areas.

With the rapid advancement of the IoT and digital manufacturing, the scale of time-series data has grown significantly [5]. At the same time, the heterogeneity and nonlinear relationships within these sequences have become increasingly complex, highlighting the limits of traditional modeling assumptions and the cost of feature engineering [6], [7].

Against this backdrop, deep learning (DL)-based methods have emerged as promising approaches that enable end-to-end learning for sequential pattern analysis. Among various applications, anomaly detection is particularly critical, as it plays a key role in the early identification of financial spikes, equipment failures, and network intrusions. Reliable detection mechanisms are essential, as they can significantly reduce potential losses [8], [9].

However, in real-world scenarios, anomalies are often rare and difficult to label, making fully supervised methods less feasible due to their reliance on complex feature engineering. In contrast, unsupervised anomaly detection approaches offer a more practical and scalable alternative [10]. These models are typically trained only on normal data, learning to model normal data behavior. During inference, when the model encounters anomalous inputs that behave significantly differently from the learned normal behavior, the resulting deviations are quantified into anomaly scores. Higher scores indicate a higher likelihood of anomalous behavior [11].

Over the past few years, large-scale pretrained models have revolutionized AI research. In the field of Natural Language Processing (NLP), BERT showed that self-supervised pretraining followed by light fine tuning can set state of the art (SOTA) results on several tasks [12]. Scaling up to GPT-3 unlocked strong task-agnostic few-shot and zero-shot abilities across translation, question answering and reasoning tasks [13]. Parallel progress in vision reveals the same trend: a pure-Transformer Vision Transformer (ViT) matches or surpasses CNNs once pre-trained on web-scale labeled images [14]. Recently, multimodal pretraining pushes the envelope further. The contrastive image-text model CLIP attains ImageNet-level accuracy without any task-specific labels [15], while LLaVA combines a vision encoder with an instruction-tuned Large Language Model (LLM) to achieve outstanding performance [16].

Taken together, these successes demonstrate that large-scale pretrained models can yield highly general, easily adapted representations. Motivated by this paradigm, we investigate whether a similarly pretrained LLM backbone, kept frozen without any fine-tuning, can serve as an effective and fully unsupervised anomaly detector for multivariate time-series tasks.

The contributions of this work are summarized as follows:

1) Proposing a Novel, Comprehensive Three-Branch LLM Architecture for Unsupervised Anomaly Detection: We propose TriP-LLM, an unsupervised multivariate time series anomaly detection framework that employs a three-branch patch-wise large language model architecture. This framework integrates local and global temporal features using a three-branch design consisting of patching, selecting, and global modules. These branches encode the input time series into a patch-wise representation, which is then processed by a frozen, pre-trained LLM. Finally, a lightweight patch-wise decoder reconstructs the input sequence to produce an anomaly score.
2) Achievement of State-of-the-Art Performance under Rigorous, Fair Evaluation: TriP-LLM demonstrates superior or competitive performance compared to existing state-of-the-art (SOTA) methods across multiple benchmark datasets. It achieves the highest average PATE score (0.5002) among all compared SOTA methods. This evaluation was conducted using PATE (Proximity-Aware Time-series anomaly Evaluation), a comprehensive, threshold-free metric that provides a more rigorous assessment than traditional F1 scores by rewarding early alarms and robust coverage. Furthermore, all baseline methods were implemented and tested under a unified open-source framework (DeepOD) to ensure fair comparison.
3) Demonstration of Critical Memory Efficiency and Deployment Scalability: The TriP-LLM framework addresses the major memory scalability bottleneck observed in conventional Channel Independence (CI)-based LLMs. TriP-LLM's triple-branch encoder design compresses the input without increasing the batch dimension, resulting in a significantly lower and nearly constant peak GPU memory footprint. Quantitatively, TriP-LLM required only 4.5 GB of peak memory compared to 44.9 GB for the CI-LLM equivalent on Gemma3-1B, making it approximately 9.9 times more memory efficient in that configuration. This memory efficiency provides a clear deployment justification, as TriP-LLM can stay within the capacity of a single consumer GPU, making it more scalable and practical for real-world deployment where CI-LLMs would significantly degrade runtime performance due to memory offloading.
4) Extensive Experimental Validation of Architectural Component Contributions: The TriP-LLM framework conducts detailed ablation studies to confirm the effectiveness and necessity of the architectural components. These studies verify that the full tri-branch structure (Patching, Selecting, and Global) is essential to robustly capture local, multi-scale, and global temporal patterns, as removing any single branch resulted in a noticeable drop in performance. Furthermore, comparisons against variants replacing the LLM (e.g., LLM2Trans or Remove LLM) confirmed that the LLM component itself contributes positively to overall performance, justifying its inclusion over simpler Transformer-based architectures.

## II. RELATED WORKS

In this section, we review recent methods and literature on multivariate time-series anomaly detection based on deep learning approaches.

### A. RECURRENT NEURAL NETWORKS

Recurrent Neural Networks (RNNs), along with their variants such as Long Short-Term Memory (LSTM) and Gated Recurrent Unit (GRU), have become widely adopted backbone models for time-series anomaly detection in recent years.

Among these, Li et al. [17] proposed a Generative Adversarial Network (GAN)-based time-series anomaly detection framework for cyber-physical systems, using LSTM generator and discriminator inside a GAN, then fuse their reconstruction and discrimination losses with a novel anomaly scoring function. Integrated with variational autoencoder (VAE), Li et al. [18] introduced a reconstruction-based hierarchical VAE detection framework that learns low-dimensional inter-channel embeddings via a stochastic RNN-like latent module, while a 1D Convolutional Neural Network (CNN) compression captures temporal dependencies.

In another line of work, Wang et al. [19] presented a prediction-based rail-transit time-series detector that uses an improved LSTM coupling the forget and input gates and injecting the current input into the output to cut false alarms in multi-device metro data. More recently, Yu et al. [20] proposed a reconstruction-based lightweight Intrusion Detection System (IDS) that utilizes the dual-GRU-AE with two simple gate networks to leverage the multi-scale features for better detection performance in the IoT traffic time-series.

### *B. TRANSFORMER MODELS*

Transformer models, due to their powerful sequence modeling capabilities, have achieved great success and have been the focus of extensive research in time-series tasks in recent years.

For example, Tuli et al. [21] proposed an adversarial-training-based Transformer framework with twin encoder-decoder blocks and a two-phase training scheme, where the first phase focuses on reconstruction, and the second adversarial phase distinguishes between the reconstructed and the real data, thereby amplifying the ability to detect anomalies.

Leveraging the association discrepancy between normal and anomalous patterns, Xu et al. [22] presented a transformer-based detection framework that contrasts the prior associations derived from a learnable Gaussian kernel with the series associations captured by vanilla self-attention; this gap is amplified through a minimax optimization strategy to better detect anomalies. Similarly, Yang et al. [23] developed a dual-attention contrastive learning framework that employs a two-branch Transformer structure to contrast patch-wise and in-patch representations from input time-series. A KL-divergence-based discrepancy score is used to evaluate anomalies.

Recently, Kang et al. [24] introduced a reconstruction-based Variable Temporal Transformer detection framework that embeds each variable stream with multi-resolution dilated-causal convolutions, then alternates variable-self-attention and temporal-self-attention modules to jointly capture inter-variable and temporal dependencies.

### *C. SEQUENCE MODELS BEYOND TRANSFORMER*

In recent years, to address the computational complexity of modeling long sequences with Transformers, numerous efficient and powerful sequence models have been proposed. For example, variants based on state space models (SSMs), such as Mamba [25], as well as improved versions of traditional LSTM architectures, such as xLSTM [26], have been explored. Several studies have applied these models to time-series anomaly detection tasks.

For instance, with the observation that normal behaviors exhibit strong correlations, Yu et al. [27] proposed a reconstruction- and representation-consistency-based IDS framework for the Internet of Medical Things (IoMT), which utilizes a bidirectional Mamba-AE with a gate mechanism to fuse forward and backward features, jointly with an MLP projector to measure latent representation discrepancy.

On the other hand, Faber et al. [28] developed a dual-mode (forecasting and reconstruction-based) encoder-decoder detection framework that stacks xLSTM blocks with exponential gating, depth-wise convolutions and residual connections to capture long-range temporal dependencies for time-series data.

### *D. LARGE LANGUAGE MODELS*

Recently, the application of LLMs across different modalities has become a highly active area of research. In the domain of time-series analysis, LLMs have demonstrated remarkable performance in tasks such as forecasting, classification, and anomaly detection.

Motivated by the success of masked language modeling in NLP, Jeong et al. [29] proposed a self-supervised, classification-style anomaly detection framework that replaces random with four kinds of synthetic outliers, and uses 1-D relative-position-biased self-attention for better anomaly detection.

In another work, Zhou et al. [30] introduced a Frozen Pretrained Transformer (FPT) based time-series analyzer by fine-tuning only lightweight projection and normalization layers of GPT-2 model, achieving state-of-the-art performance across time-series classification, anomaly-detection, short-term and long-term forecasting, imputation, and few-/zero-shot settings.

Similarly, Bian et al. [31] developed an LLM-based framework that reframes time-series forecasting as a self-supervised multi-patch prediction problem and replaces the heavy, overfitting-prone sequence-level head with a lightweight shared patch-wise decoder. A two-stage scheme, consisting of causal next-patch continual pre-training followed by multi-patch prediction fine-tuning, delivers strong performance on downstream tasks such as anomaly detection, forecasting, imputation, and classification.

To further enhance the alignment between time-series and language models, Jin et al. [32] designed an LLM-based framework for time-series forecasting that reprograms time-series patches to align with the modalities of natural language via multi-head cross-attention and leverages Prompt-as-Prefix instructions to steer a frozen LLM backbone toward accurate prediction.

Leveraging the inherent autoregressive generation ability of a frozen decoder-only LLM, Liu et al. [33] proposed an autoregressive LLM-based forecaster that projects time-series segments into the model's token space via trainable segment and timestamp embeddings, then iteratively predicts future segments via next-token generation while also supporting time-series-as-prompt in-context forecasting.

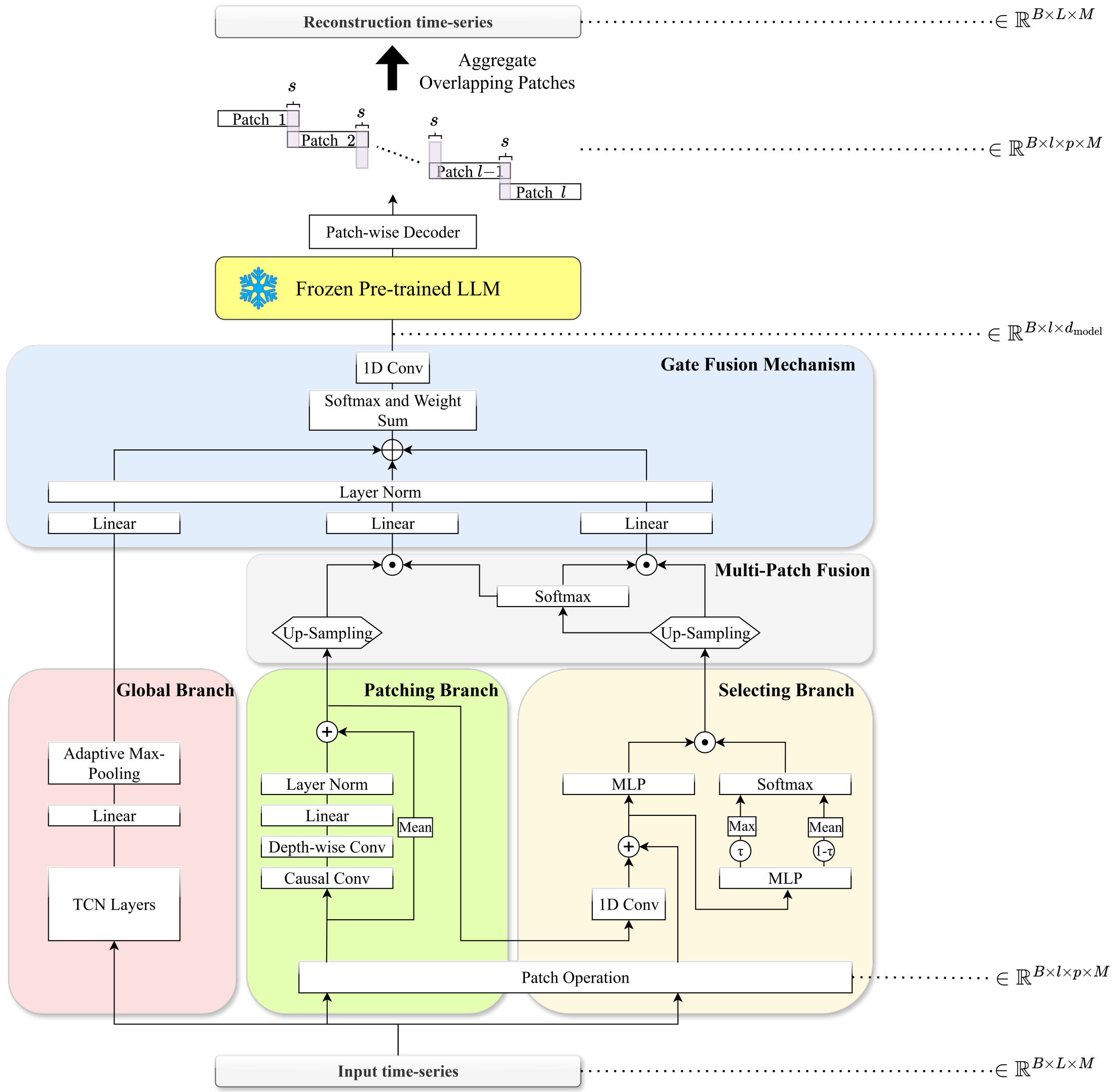

FIGURE 1. Overview of the TriP-LLM framework. The input multivariate time-series is processed through three specialized branches, namely Patching, Selecting, and Global, to extract local and global temporal features. These features are fused via a gate-fusion mechanism and transformed into patch-wise representations, which are passed into a frozen pretrained LLM. The output is then decoded by a patch-wise decoder to reconstruct the input sequence and compute anomaly scores.

In the zero-shot setting, some approaches directly feed time-series data into LLMs without task-specific fine-tuning. These methods have been investigated for tasks such as forecasting and anomaly detection. For example, Gruver et al. [34] proposed a zero-shot, tokenization-based forecasting framework that encodes numerical time-series values as digit strings, frames forecasting as next-token prediction in a frozen LLM, and converts the model's discrete token probabilities into flexible continuous densities, thus demonstrating the capability of zero-shot LLM forecasters.

Another line of work, Alnegheimish et al. [35] designed an LLM-based zero-shot detection framework that first converts scaled time-series windows into digit-token strings, then detects anomalies through two frozen LLM pipelines: PROMPTER, which flags outlier tokens via prompting, and DETECTOR, which forecasts the next values.

## III. PRELIMINARIES

We define a multivariate time-series $[x_1, x_2, \ldots, x_L] \in \mathbb{R}^{L \times M}$, where each $x_t \in \mathbb{R}^M$ is a measurement vector with $M$ channels at time point $t$, and $L$ denotes the total sequence length. When $M = 1$, the series reduces to a univariate case.

In the context of semi-supervised or unsupervised anomaly detection approaches, it is commonly assumed that the training set consists entirely of normal behavior samples. The goal is to determine whether each time point in the test sequence exhibits anomalous behavior, typically by assigning a binary label based on a predefined threshold (e.g., 0 for normal, 1 for anomalous).

## IV. METHOD

Fig. 1 illustrates the overall architecture of TriP-LLM, which transforms input multivariate time-series into a patch-wise representations through three dedicated branches.

First, the Patching Branch segments the input sequence into overlapping patches with patch size $p$ and stride $s$. The second branch, Selecting Branch, further filters these local patches to highlight informative temporal segments. The third, the Global Branch, captures long-range temporal dependencies across the entire sequence.

These three branches are then fused via a gate-modulated fusion module to produce the final patch-wise input sequence. This sequence is then passed to a frozen pretrained LLM backbone without any finetuning. The LLM's output, a sequence of patch-level embeddings, is finally decoded patch-by-patch through a shared patch-wise decoder to reconstruct the original input.

### A. PATCHING BRANCH

To effectively capture the local temporal dynamics of input multivariate time-series, we introduce a two-stage processing module composed of a Patching Branch and a Selecting Branch, both operating on patch-wise representations of the input.

The Patching Branch is responsible for extracting local contextual features from the input time-series using a patching and causal convolutional processing pipeline. First, we consider a batch of multivariate time-series $\boldsymbol{X} \in \mathbb{R}^{B\times L\times M}$, where $B$ is the batch size, $L$ is the time-series length, $M$ denotes the number of the channels. We first segment the sequence into overlapping patches of size $p$ with stride $s$, resulting in a patch tensor $\boldsymbol{X}_{\boldsymbol{patch}}$:

$$\boldsymbol{X}_{\boldsymbol{patch}} \in \mathbb{R}^{B\times l\times p\times M}, \tag{1}$$

where $l = \left\lfloor \frac{L-p}{s} \right\rfloor + 1$ is the number of patches generated.

Each patch is then processed by a two-layer causal convolutional $\mathrm{Conv}_{\mathrm{causal}}$ network with increasing dilation rates, enabling receptive field expansion over time without leakage from the future to align with the nature of time-series.

Next, a depth-wise convolution $\mathrm{Conv}_{\mathrm{dw}}$ is applied to each channel independently, followed by a linear projection to the dimension $d$, with residual connection from the patch mean, allowing the model to preserve sliding patch features $\boldsymbol{\mathcal{F}_p}$, as shown in the equation below:

$$\boldsymbol{\mathcal{F}_p} = \mathrm{LayerNorm}\left(\mathrm{Linear}\left(\mathrm{Conv}_{\mathrm{causal,\ dw}}\left(\boldsymbol{X}_{\boldsymbol{patch}}\right)\right)\right) + \mathrm{Mean}(\boldsymbol{X}_{\boldsymbol{patch}}). \tag{2}$$

Finally, we reshape the representation $\boldsymbol{\mathcal{F}_p}$ back into shape $\in \mathbb{R}^{B\times l\times (M\cdot d)}$, forming a local patch-based token embedding.

### B. SELECTING BRANCH

To identify and emphasize semantically important patches, we introduce a Selecting Branch based on patch weighting. The Selecting Branch operates in two stages: scoring and feature transformation.

First, each patch is first combined with the output of the Patching Branch via additive modulation with 1D convolution $\mathrm{Conv}_{\mathrm{1D}}$ to align the tensor shape:

$$\widetilde{\boldsymbol{\mathcal{F}_p}} = \boldsymbol{X}_{\boldsymbol{patch}} + \mathrm{Conv}_{\mathrm{1D}}(\boldsymbol{\mathcal{F}_p}), \tag{3}$$

We then compute importance score $s_i$ for each patch $i$ by passing $\widetilde{\boldsymbol{\mathcal{F}_p}}$ into the small MLP-based network. To ensure a stable and expressive weighting score $\tilde{s}_i$, we use a learnable parameter $\tau$ to fuse max- and mean-pooling across channels $M$:

$$\tilde{s}_i = \tau \cdot \mathrm{MaxPooling}(s_i, M) + (1-\tau)\cdot \mathrm{MeanPooling}(s_i, M). \tag{4}$$

Then, the weights over patches are computed by Softmax function:

$$\boldsymbol{\alpha} = \mathrm{Softmax}(\tilde{s}). \tag{5}$$

In parallel, $\widetilde{\boldsymbol{\mathcal{F}_p}}$ is projected to the dimension $d$ with small MLP network. The final selected representation $\boldsymbol{\mathcal{F}_{sle}}$ is computed by elementwise multiplication between $\boldsymbol{\alpha}$ and projected $\widetilde{\boldsymbol{\mathcal{F}_p}}$:

$$\boldsymbol{\mathcal{F}_{sle}} = \boldsymbol{\alpha} \odot \mathrm{Linear}(\widetilde{\boldsymbol{\mathcal{F}_p}}). \tag{6}$$

This results in a representation $\in \mathbb{R}^{B\times l\times (M\cdot d)}$, aligned in shape with the Patching Branch.

### C. MULTI-PATCH FUSION

To enhance the robustness and expressiveness of local representations, we employ a multi-scale patching strategy. Specifically, we instantiate the above two branches with different $S$ patch sizes $\{p_1, p_2, \ldots, p_S\}$. For each patch size $k$ and its corresponding segmented number of patches $l_k$, we obtain two outputs:

1. $\boldsymbol{\mathcal{F}}_p^k \in \mathbb{R}^{B\times l_k\times (M\cdot d)}$ obtained from the Patching Branch.
2. $\boldsymbol{\mathcal{F}}_{sle}^k \in \mathbb{R}^{B\times l_k\times (M\cdot d)}$ obtained from the Selecting Branch.

Since smaller $k$ would lead to more $l_k$ than those of larger $k$, to enable fusion across different patch resolutions, we upsample all features to a maximum patch length $l_{max}$ using 1D interpolation, generating the up-sampled feature $\widetilde{\boldsymbol{\mathcal{F}}}^k$:

$$\widetilde{\boldsymbol{\mathcal{F}}}^k = \mathrm{Upsample}(\boldsymbol{\mathcal{F}}^k, l_{max}). \tag{7}$$

Utilizing the mean of the aligned value of the multi-patch Selecting Branch, a patch-scale-wise Softmax is then applied to fuse each branch output, yield the final representations $\widehat{\mathcal{F}}_p$ and $\widehat{\mathcal{F}}_{sle}$, capturing patch-wise features from the multi-patch-scale, described as follows:

$$\widehat{\mathcal{F}}_p = \sum_{k=1}^{S} \text{Softmax}(\text{Mean}(\widetilde{\mathcal{F}}_{sle}^k)) \cdot \widetilde{\mathcal{F}}_p^k, \quad (8)$$

$$\widehat{\mathcal{F}}_{sle} = \sum_{k=1}^{S} \text{Softmax}(\text{Mean}(\widetilde{\mathcal{F}}_{sle}^k)) \cdot \widetilde{\mathcal{F}}_{sle}^k. \quad (9)$$

### D. GLOBAL BRANCH

To complement the local patch-wise representations, we incorporate a Global Branch designed to model long-range temporal dependencies across the entire input sequence. We pass the original batch of multivariate time-series $\boldsymbol{X} \in \mathbb{R}^{B \times L \times M}$ into a stacked temporal convolutional network (TCN) with increasing receptive fields. Then, the output is then projected into dimension $d$, followed by an adaptive max-pooling to match the temporal length $l_{max}$ of other local patch-based branches, generating the global representation $\mathcal{F}_g$ :

$$\mathcal{F}_g = \text{AdaptMaxPooling}(\text{Linear}(\text{TCN}(\boldsymbol{X}))). \quad (10)$$

This global feature $\mathcal{F}_g \in \mathbb{R}^{B \times l \times d}$ is time-aligned and dimension-matched with the outputs from the local Patching and Selecting branches.

### E. GATE-FUSION MECHANISM

To effectively integrate the three feature branches, we introduce a gate-fusion mechanism. Each branch is first projected into a unified latent dimension $D'$ with shared layer normalization operation:

$$\boldsymbol{P}' = \text{LayerNorm}\left(\text{Linear}\left(\widehat{\mathcal{F}}_p\right)\right), \quad (11)$$

$$\boldsymbol{S}' = \text{LayerNorm}\left(\text{Linear}\left(\widehat{\mathcal{F}}_{sle}\right)\right), \text{ and} \quad (12)$$

$$\boldsymbol{G}' = \text{LayerNorm}\left(\text{Linear}\left(\mathcal{F}_g\right)\right). \quad (13)$$

Then, we concatenate the three projected tokens at each time point and compute gate weights $\boldsymbol{\beta} \in \mathbb{R}^{B \times l_{max} \times 3}$ via a linear layer:

$$\boldsymbol{\beta} = \text{Softmax}(\text{Linear}([\boldsymbol{P}', \boldsymbol{S}', \boldsymbol{G}'])). \quad (14)$$

Let the weights be split as $\boldsymbol{\beta_P}$, $\boldsymbol{\beta_S}$ and $\boldsymbol{\beta_G}$ to denote the weight for corresponding branch. The fused feature representation $\mathcal{F}_{fuse}$ is obtained by a weighted sum:

$$\mathcal{F}_{fuse} = \boldsymbol{\beta_P} \odot \boldsymbol{P}' + \boldsymbol{\beta_S} \odot \boldsymbol{S}' + \boldsymbol{\beta_G} \odot \boldsymbol{G}'. \quad (15)$$

TABLE I
SUMMARY STATISTICS OF DATASETS

| Datasets | Training Samples | Test Samples | Anomaly Ratios | Channels |
|---|---|---|---|---|
| SMD | 708,405 | 708,420 | 4.16 % | 38 |
| SWAT | 495,000 | 449,919 | 12.14 % | 51 |
| MSL | 58,317 | 73,729 | 10.53 % | 55 |
| PSM | 132,481 | 87,841 | 27.76 % | 25 |
| NIPS-TS-SWAN | 60,000 | 60,000 | 32.60 % | 38 |

Finally, we apply a 1D convolution to project the fused representation into the LLM input space with dimension $d_{model}$, resulting in a patch-wise embedding $\in \mathbb{R}^{B \times l_{max} \times d_{model}}$ that integrates both global and local multi-scale temporal patterns.

### F. PATCH-WISE DECODER

To reconstruct the original multivariate time-series, we employ a lightweight patch-wise decoder that operates on the output of the LLM in a token-by-token manner.

Instead of generating the entire sequence at once, our decoder processes each token, corresponding to a specific patch, individually using a shared MLP. Each token is decoded into a fixed-length patch, and overlapping regions are merged via average pooling to reconstruct the full sequence. This design not only simplifies the decoding process but also naturally aligns with our patch-based LLM input format.

Notably, this shared patch-wise decoder architecture has been shown in prior work to be more parameter-efficient and less prone to overfitting than a full sequence-level flattened decoder [31].

## V. EXPERIMENT

### A. DATASET

We evaluate our approach on five widely used multivariate time-series anomaly detection benchmarks: SMD [36], SWaT [37], MSL [38], PSM [39], and NIPS-TS-SWAN [40], [41]. These datasets span a diverse range of real-world domains, including server monitoring, cyber-physical systems, and telemetry. Each dataset exhibits unique characteristics in terms of dimensionality and anomaly types, providing a comprehensive testbed for unsupervised anomaly detection. Detailed dataset statistics and properties are summarized in Table I.

### B. EVALUATION METRIC

In recent years, although the traditional F1 score has been widely used as an evaluation metric, its point-wise nature that focuses only on individual times points, often underestimates the capability of time-series anomaly detection models [11].

TABLE II
COMPARISON OF PATE SCORES ACROSS BENCHMARK DATASETS

| Datasets<br>Methods | SMD | SWaT | MSL | PSM | NIPS-TS-SWAN | AVG |
|---|---|---|---|---|---|---|
| USAD | 0.1683 | 0.7292 | 0.1673 | 0.4584 | 0.6531 | 0.4353 |
| TranAD | 0.1388 | 0.7287 | 0.1641 | 0.4457 | 0.6745 | 0.4304 |
| AnomTrans | 0.2084 | 0.7159 | 0.1712 | 0.5220 | 0.6683 | 0.4572 |
| TimesNet | 0.1950 | 0.1171 | 0.1737 | 0.4285 | 0.6732 | 0.3175 |
| DIF | 0.2289 | 0.7178 | 0.1786 | 0.4978 | 0.7188 | 0.4684 |
| DCdetector | 0.0579 | 0.1186 | 0.1186 | 0.2982 | 0.4615 | 0.2110 |
| PatchAD | 0.0573 | 0.1222 | 0.1173 | 0.2985 | 0.4818 | 0.2154 |
| GPT4TS | 0.1840 | 0.0944 | 0.1827 | 0.4853 | 0.6570 | 0.3207 |
| Time-LLM | 0.2272 | 0.0898 | 0.1924 | 0.4884 | 0.6157 | 0.3227 |
| CBMAD | 0.2125 | 0.6948 | 0.1911 | 0.5495 | 0.7383 | 0.4772 |
| TriP-LLM | **0.2411** | **0.7352** | **0.2146** | **0.5671** | **0.7431** | **0.5002** |

[a] The highest value is shown in **red**; the second highest is underlined.

TABLE III
PATE SCORES OF TRIP-LLM USING DIFFERENT FROZEN LLM BACKBONES

| Datasets<br>Methods | SMD | SWaT | MSL | PSM | NIPS-TS-SWAN | AVG |
|---|---|---|---|---|---|---|
| LLaMA3.2-1B | 0.1981 | **0.7503** | 0.2028 | 0.4881 | **0.7133** | 0.4705 |
| Gemma3-1B | 0.1959 | 0.7122 | **0.2043** | 0.5034 | 0.7036 | 0.4639 |
| Qwen3-0.6B | **0.2321** | 0.7216 | 0.2028 | **0.5592** | 0.7019 | **0.4835** |

[a] The highest value is shown in **red**.

To address this, several evaluation metrics specifically designed for time-series settings have been proposed, such as the composite F1-score that combines point-wise and event-based assessment [11], and affiliation metrics that account for both temporal continuity and detection offset [42].

On the other hand, the choice of thresholding strategy has a significant impact on detection performance. For instance, the best-F1 threshold approach exhaustively searches for all possible thresholds that maximize the F1 score, thereby estimating the upper bound of a model's detection ability.

However, for more rigorous and fair comparisons across different methods, we require an evaluation metric that considers performance across all settings, rather than relying on a single fixed value. This not only reduces bias introduced by an optimally tuned threshold but also reveals a detection model's robustness across operating conditions.

To obtain fair and robust experimental comparisons, we employ **PATE** (Proximity-Aware Time-series anomaly Evaluation) [43], a recently proposed threshold-free metric tailored for time-series anomaly detection. PATE departs from classical area-under-the-curve metrics such as AUC-PR and AUC-ROC in two key respects. (i) Rather than treating every time point equally, it assigns proximity-based weights to predictions, rewarding early or on-time alarms while progressively discounting late or distant ones. (ii) It evaluates performance across a grid of pre- and post-buffer lengths and over the full threshold range, then averages the resulting weighted AUC-PR values, yielding a single scalar score with no extra hyper-parameters. This formulation simultaneously captures early-warning ability, coverage of the entire anomaly span, and robustness to threshold choice. These are qualities that conventional AUC or even the volume-under-the-surface (VUS) metrics [44] cannot fully reflect because they lack proximity-aware weighting. Consequently, PATE offers a stricter yet fairer assessment of real-world detection quality.

### C. EVALUATION

Based on the PyTorch framework, we performed model training and inference on an AMD R9 7900X CPU and a single NVIDIA RTX 4090 24GB GPU. We adopted the officially released pretrained GPT-2 [45] model as the LLM backbone in our TriP-LLM framework and kept the LLM frozen during training, with no gradient updates or optimization applied. Detailed training hyperparameters and model checkpoints are available in our publicly released code on GitHub.

We compared our method against several recent SOTA methods known for their strong performance in time-series anomaly detection tasks, including USAD [46], TranAD [21], AnomTrans [22], TimesNet [47], DIF [48], DCdetector [23], PatchAD [49], and the Mamba-based method CBMAD [27].

In addition, we compared with other LLM-based methods for time-series tasks, such as GPT4TS [30] and Time-LLM [32]. Because Time-LLM was originally designed for forecasting, to adapt it to the anomaly detection task, we made necessary modifications, including crafting description

TABLE IV
PATE SCORES FROM THE ABLATION STUDY OF TRIP-LLM

| Methods \ Datasets | SMD | SWaT | MSL | PSM | NIPS-TS-SWAN | AVG |
|---|---|---|---|---|---|---|
| TriP-LLM | **0.2411** | **0.7352** | **0.2146** | **0.5671** | **0.7431** | **0.5002** |
| w/o Selection | 0.2190 | 0.2553 | 0.2024 | 0.5159 | 0.7104 | 0.3806 |
| w/o Patching | 0.2361 | 0.7077 | 0.2012 | 0.5618 | 0.7247 | 0.4863 |
| w/o Global | 0.2384 | 0.1442 | 0.2031 | 0.5638 | 0.7197 | 0.3738 |
| Base LLM | 0.2180 | 0.7141 | 0.2046 | 0.5633 | 0.7407 | 0.4881 |
| Seq-decoder | 0.2281 | 0.1277 | 0.2021 | 0.5399 | 0.7325 | 0.3661 |
| Remove LLM | 0.2169 | 0.6104 | 0.2145 | 0.5511 | 0.7121 | 0.4610 |
| LLM2Trans | 0.2091 | 0.7325 | 0.2133 | 0.4803 | 0.7104 | 0.4691 |
| LLM2Atten | 0.2101 | 0.4648 | 0.2037 | 0.5116 | 0.7353 | 0.4251 |

a. The highest value is shown in **red**; the second highest is underlined.

TABLE V
AVERAGE PEAK GPU MEMORY (GB) IN INFERENCE MODE FOR TRIP-LLM VS. CI-LLM ACROSS BACKBONE MODELS, BATCH SIZES, PATCH SIZES, AND CHANNEL DIMENSIONS

| Methods [patch size] | Batch size 2 | | | Batch size 4 | | | Batch size 8 | | | Batch size 16 | | |
|---|---|---|---|---|---|---|---|---|---|---|---|---|
| **Channel Dimensions** | **25** | **51** | **55** | **25** | **51** | **55** | **25** | **51** | **55** | **25** | **51** | **55** |
| GPT2 CI [8] | 0.88 | 1.30 | 1.36 | 1.28 | 2.11 | 2.24 | 2.08 | 3.74 | 4.00 | 3.67 | 6.99 | 7.51 |
| GPT2 TriP [8] | 0.52 | 0.52 | 0.52 | 0.53 | 0.54 | 0.54 | 0.56 | 0.58 | 0.58 | 0.63 | 0.66 | 0.66 |
| GPT2 CI [16] | 0.81 | 1.14 | 1.19 | 1.13 | 1.79 | 1.91 | 1.77 | 3.11 | 3.31 | 3.05 | 5.73 | 6.14 |
| GPT2 TriP [16] | 0.52 | 0.52 | 0.52 | 0.52 | 0.53 | 0.53 | 0.55 | 0.56 | 0.56 | 0.61 | 0.63 | 0.63 |
| Llama3.2-1B CI [8] | 5.98 | 7.41 | 7.63 | 7.35 | 10.21 | 10.64 | 10.09 | 15.80 | 16.67 | 15.58 | 26.99 | 28.74 |
| Llama3.2-1B TriP [8] | 4.68 | 4.69 | 4.69 | 4.73 | 4.75 | 4.75 | 4.85 | 4.86 | 4.86 | 5.08 | 5.10 | 5.10 |
| Llama3.2-1B CI [16] | 5.72 | 6.86 | 7.04 | 6.82 | 9.12 | 9.47 | 9.03 | 13.62 | 14.32 | 13.44 | 22.62 | 24.03 |
| Llama3.2-1B TriP [16] | 4.67 | 4.67 | 4.68 | 4.71 | 4.72 | 4.72 | 4.80 | 4.82 | 4.82 | 4.99 | 5.01 | 5.01 |
| Gemma3-1B CI [8] | 6.07 | 8.51 | 8.88 | 8.42 | 13.29 | 14.04 | 13.10 | 22.84 | 24.34 | 22.46 | 41.94 | 44.94 |
| Gemma3-1B TriP [8] | 3.84 | 3.85 | 3.85 | 3.93 | 3.94 | 3.94 | 4.12 | 4.14 | 4.14 | 4.51 | 4.53 | 4.53 |
| Gemma3-1B CI [16] | 5.63 | 7.58 | 7.88 | 7.50 | 11.42 | 12.03 | 11.27 | 19.11 | 20.32 | 18.81 | 34.49 | 36.90 |
| Gemma3-1B TriP [16] | 3.82 | 3.83 | 3.83 | 3.90 | 3.90 | 3.91 | 4.05 | 4.06 | 4.06 | 4.36 | 4.38 | 4.38 |

prompts specific to each dataset. Furthermore, due to GPU memory limitations that prevented us from running the LLaMA-7B version of Time-LLM, we replaced it with GPT-2 as a substitute.

To ensure a fair comparison, we implemented all baseline methods, including TriP-LLM, under the open-source and widely used time-series anomaly detection framework DeepOD [48], [50].

As shown in Table II, the overall performance demonstrates that our proposed TriP-LLM achieves the best detection performance under the comprehensive PATE metric, even across datasets collected from diverse domains with varying anomaly ratios.

Moreover, we validated TriP-LLM using larger small-scale LLMs, including LLaMA3.2-1B [51], Gemma3-1B [52], and Qwen3-0.6B [53]. All models were evaluated without adjustment to the original training hyperparameters used for the GPT-2 version of TriP-LLM. As shown in Table III, these models still delivered strong performance, with average PATE scores surpassing almost all baseline methods. This demonstrates the robustness and generalizability of the proposed TriP-LLM.

### D. ABLATION STUDY

To evaluate the contribution of each individual component in our proposed model, we conducted a comprehensive ablation study on various variants of TriP-LLM.

We first investigated the impact of each of the three input branches. The variant **w/o Selection** removes the Selection Branch, meaning the model is unable to selectively process the patch sequences and thus cannot emphasize the most informative local temporal segments. The **w/o Patching** variant removes the Patching Branch, resulting in a loss of the model's capability to capture local temporal patterns. In the **w/o Global** setting, the Global Branch is omitted, which restricts the model from modeling long-range temporal dependencies and limits it to local temporal features only.

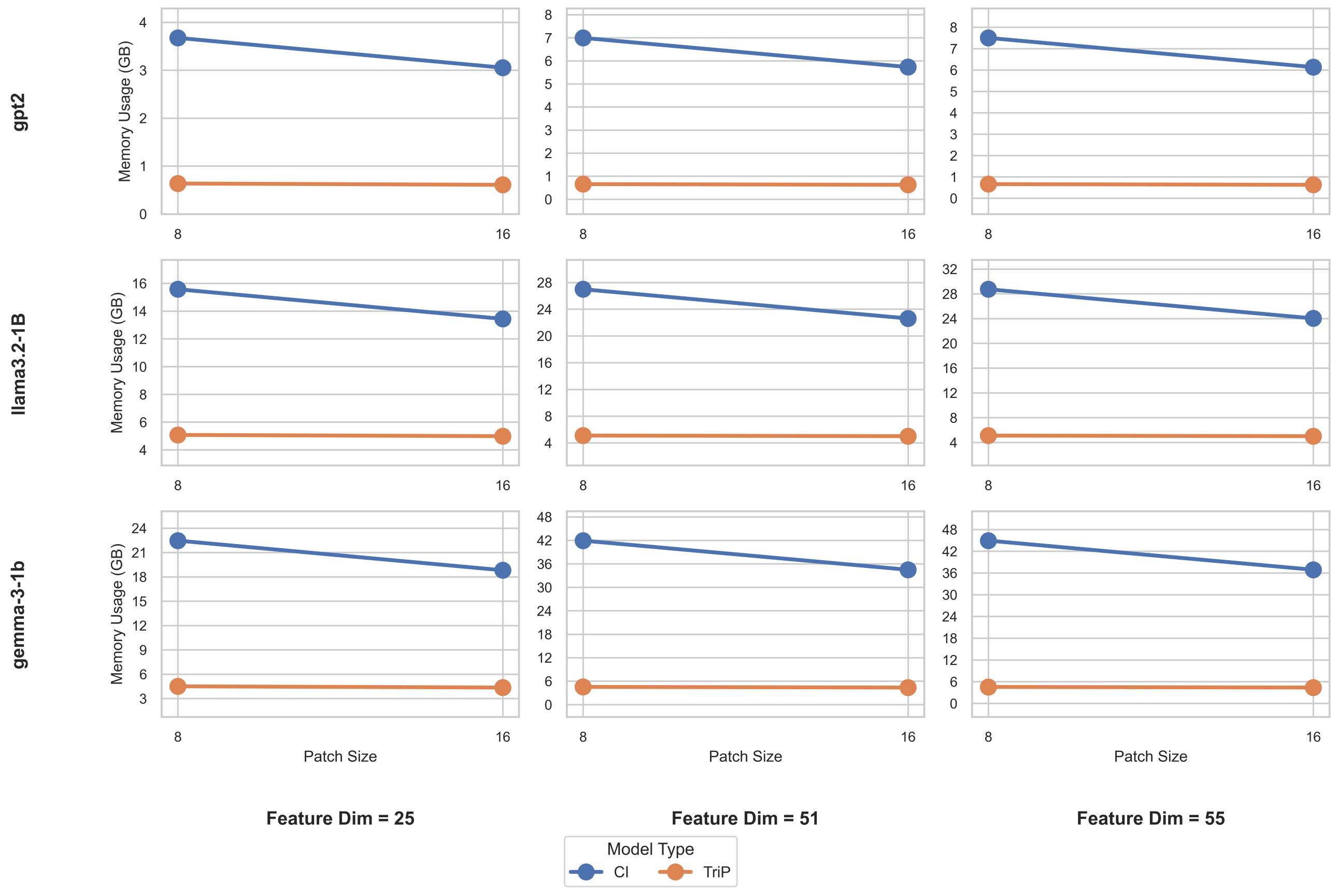

**FIGURE 2. Peak GPU memory usage (GB) of TriP-LLM vs. CI-LLM (batch size = 16, inference mode).**

Additionally, we examined a minimal version of the model in which all three branches are removed and a single linear layer is used to project the input directly to the output. We refer to this variant as **Base LLM**. This configuration is designed to assess whether the architectural branches design for the LLM contributes meaningfully to the overall detection performance.

We also ablated the output processing mechanism of the TriP-LLM. While prior work [31] has demonstrated the advantages of a patch-wise decoder over a heavy flattened-head decoder, we introduced a variant called **Seq-decoder**, which replace the patch-wise decoder with a flattened-head decoder to validate this design choice in our context.

Importantly, recent studies have questioned the effectiveness of LLMs in time-series tasks, arguing that their contribution may be negligible. In response, we incorporated three additional ablation variants proposed by this line of research [54] to re-examine the role of the LLM in our model. The first, **Remove LLM**, eliminates the LLM entirely. The second, **LLM2Trans**, replaces the LLM with a Transformer encoder. The third, **LLM2Atten**, substitutes the LLM with a multi-head attention module.

As shown in Table IV, the results of all the ablation experiment collectively demonstrates that each component, including the input tri-branch design, the use of the LLM itself, and the output decoding strategy, contributes positively to the model's overall anomaly detection performance. These findings further validate the effectiveness of the proposed TriP-LLM framework.

### E. MOTIVATION FOR THE TRI-BRANCH ENCODING

The motivation behind encoding the input multivariate time-series into a representation of shape $\in \mathbb{R}^{B\times l\times d_{model}}$, using a triple-branch structure stems from empirical observations in our experiments.

The overlapping patch operation has been widely adopted in time-series analysis as an effective technique to capture local patterns while simultaneously reducing the input sequence length. Given an input time sequence $\boldsymbol{X} \in \mathbb{R}^{B\times L\times M}$, the patching operation segments it into overlapping patch sequences $\boldsymbol{X_{patch}} \in \mathbb{R}^{B\times l\times p\times M}$. To enable the model to better capture temporal dynamics within these local patches in a single channel, the **Channel Independence (CI)** strategy is often applied. Specifically, the channel dimension $M$ is flattened into the batch dimension, resulting in a reshaped input of $(B \cdot M) \times l \times p$, which is then projected and passed to the model backbone. This approach has been demonstrated to be effective in numerous SOTA works across a variety of time-series tasks [55], [56], [57] including those using LLM-based methods [31], [32].

Despite the well-documented gains in efficiency and convergence speed of CI; however, in our experiments, we observed a significant limitation of this CI-based approach when used in conjunction with LLMs: CI-based LLMs use significantly higher GPU memory consumption, even when using relatively small LLMs. This suggests that the challenge is not in the effectiveness of CI-based LLM methods, but in the practical challenge with respect to memory scalability.

To evaluate this concern, we built CI-LLM, which directly projects and feeds the CI patch sequences into the LLM backbone. We compared the GPU memory usage between TriP-LLM and CI-LLM. For a fair comparison, we removed the decoder modules from both models and measured only the peak GPU memory allocated during a forward pass in **torch.inference_mode()** (i.e., inference-only, without gradients). All the LLMs were implemented using Hugging Face Transformers with the eager attention implementation. For each configuration we measured **torch.cuda.max_memory_allocated()** after resetting peak statistics, and repeated each measurement 200 times; values reported in Table V are the arithmetic mean (in GB) across runs. Experiments used input sequence length of 48, a patch stride of 1, and floating-point 32 (FP32) precision, we compared the peak GPU memory consumption across different models, including GPT-2, LLaMA3.2-1B, and Gemma3-1B. The evaluation was conducted under varying batch sizes (2, 4, 8, 16), patch sizes (8, 16), and across datasets with different channel dimensionalities, such as PSM (25 channels), SWaT (51 channels), and MSL (55 channels). Implementation details, such as model loading and device placement, may cause some initialization or weight-transfer overheads to be included in the measured peak. Reported numbers should therefore be regarded as conservative upper bounds. Table V summarizes the averages, and Fig. 2 shows selected results.

Across all backbones and channel counts, TriP-LLM exhibits a nearly constant memory footprint, requiring approximately 0.5–0.7 GB for GPT-2 and approximately 3.8–5.1 GB for the 1B-class models. This stability stems from the triple-branch encoder design, which compresses input time-series into patch-wise tokens without increasing the batch dimension. In contrast, CI-LLM's memory consumption grows roughly linearly with $B \times M$. For example, with batch size 16 and channel dimension 55, CI-LLM on Gemma3-1B reaches a peak allocation of 44.9 GB, compared with 4.5 GB for TriP-LLM on the same configuration (about 9.9 times larger); for the corresponding GPT-2 configuration the peak ratio is about 11.4 times. This value exceeds the 24 GB of physical VRAM available on an RTX 4090, causing part of the memory to be offloaded to system RAM via CUDA Unified Memory, which significantly degrades runtime performance. This confirms that the bottleneck of CI-based LLMs is not algorithmic effectiveness but memory scalability, whereas TriP-LLM stays within the capacity of a single consumer GPU.

Due to hardware constraints, we were unable to include larger LLMs or batch size in our experiments. However, the current results clearly demonstrate that TriP-LLM not only maintains strong anomaly detection performance, but also offers a memory-efficient and hardware-friendly alternative for time-series modeling with LLMs.

## VI. CONCLUSION

In this work, we propose TriP-LLM, a triple-branch encoder architecture that integrates both local and global temporal features from multivariate time-series inputs. The model encodes the input into a patch-wise representation, which is then fed into a frozen pretrained LLM, followed by a patch-based decoder for reconstruction.

Experimental results demonstrate that TriP-LLM achieves strong anomaly detection performance across multiple benchmark datasets. Beyond validating the model's effectiveness, we further conduct an in-depth analysis comparing TriP-LLM with the commonly used CI-based LLMs for time-series modeling. Our findings highlight a critical advantage of TriP-LLM in memory efficiency: it consistently requires significantly less GPU memory, making it more scalable and practical for real-world deployment, even with 1B-scale LLMs.

In future work, we plan to extend TriP-LLM to support larger backbone models and investigate fine-tuning strategies to further boost accuracy while maintaining the memory advantage, even online anomaly detection in streaming settings.

A preliminary version of this work was posted as a preprint on arXiv [58].

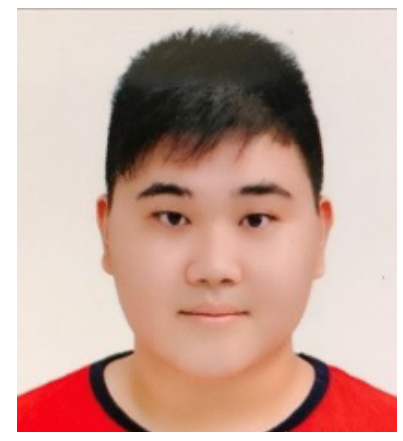

**Yuan-Cheng Yu** received his B.S. degree in electrical engineering from National Chung Hsing University, Taiwan, in 2023. He is currently pursuing the M.S. degree in the Department of Electrical Engineering at National Chung Hsing University, Taiwan. His research interests mainly include Internet of Things security, network security, deep learning, and machine learning.

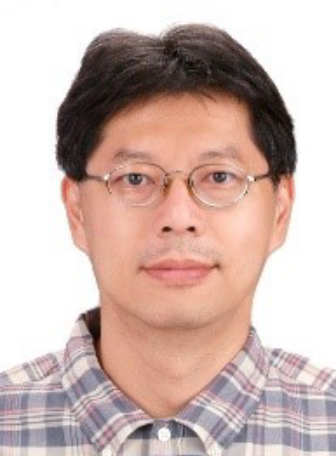

**Yen-Chieh Ouyang** (Life Member, IEEE) received his Bachelor of Science degree in Electrical Engineering from Feng Chia University, Taiwan, in 1981. He then pursued further studies in the United States, earning a Master of Science degree in Electrical Engineering (1987) and a Doctor of Philosophy degree in Electrical Engineering (1992) from the University of Memphis, Memphis, TN, USA. In August 1992, Dr. Ouyang joined the Department of Electrical Engineering at National Chung Hsing University, Taiwan, where he is currently a Professor. His research expertise lies in information-theoretical image processing with applications in medical imaging and remote sensing. He has significant experience in developing hyperspectral imaging models for Magnetic Resonance (MR) imaging applications and in the software engineering aspects of implementing and testing these methods. Furthermore, his research interests extend to the development and evaluation of cloud computing systems for various real-world applications. He also integrates artificial intelligence, machine learning, cognitive science, and deep neural networks into his work, applying these to intelligent systems such as decision support systems, human-computer interaction, and machine learning, as well as to algorithmic problems aimed at improving efficiency. His diverse research interests include: Hyperspectral image processing, Medical image processing, Agricultural production applications, Communication networks, Network security in mobile networks, Cloud computing, Multimedia system design and performance evaluation.

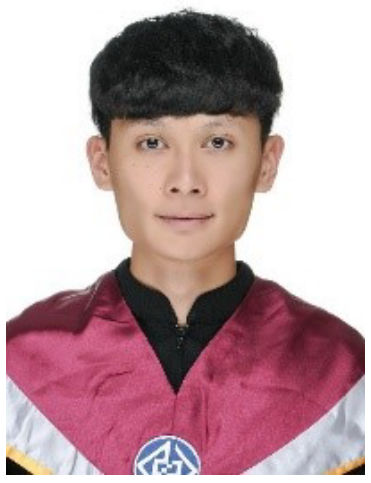

**Chun-An Lin** received his B.S. degree in Computer Science & Communication Engineering from Providence University in 2019 and his M.S. degree in Multimedia Design from National Taichung University of Science and Technology in 2021. He is currently a Ph.D. Candidate in the Department of Electrical Engineering at National Chung Hsing University. His research focuses on Machine Learning, Deep Learning, Network Security, Medical Image Analysis, and Image Super-Resolution.